\documentclass[10pt]{llncs}
\usepackage{times}
\usepackage{graphicx}
\usepackage{latexsym}
\usepackage{epic}
\usepackage{amsmath}
\usepackage{amssymb}
\usepackage{amsmath}

\input{epsf}

\begin{document}
\newcommand{\mydiv}{\mbox{\rm div}}
\newcommand{\mymod}{\mbox{\scriptsize \rm \ mod \ }}
\newcommand{\mymin}{\mbox{\rm min}}
\newcommand{\mymax}{\mbox{\rm max}}
\newcommand{\calC}{{\cal C}}
\newcommand{\calA}{{\cal A}}
\newcommand{\calL}{{\cal L}}
\newcommand{\dom}{{dom}}
\newcommand{\real}{\ensuremath{\mathbb{R}}}
\newcommand{\nat}{\ensuremath{\mathbb{N}}}
\newcommand{\ZZ}{\ensuremath{\mathbb{Z}}}

\newcommand{\set}{\mathcal}
\newcommand{\myset}[1]{\ensuremath{\mathcal #1}}
\newcommand{\myomit}[1]{}
\newcommand{\tighter}{\mbox{$\preceq$}}
\newcommand{\stighter}{\mbox{$\prec$}}
\newcommand{\incomparable}{\mbox{$\bowtie$}}
\newcommand{\equivalent}{\mbox{$\equiv$}}

\newcommand{\reg}{\mbox{$RE$}}
\newcommand{\mreg}{\mbox{$MR$}}
\newcommand{\hprs}{\mbox{$HPRS$}}
\newcommand{\reglo}{\mbox{$LO$}}
\newcommand{\ps}{\mbox{$PS$}}
\newcommand{\tc}{\mbox{$ST$}}
\newcommand{\among}{\mbox{$AD$}}
\newcommand{\lse}{\mbox{$LG$}}
\newcommand{\lser}{\mbox{$LG_R$}}
\newcommand{\cs}{\mbox{$CS$}}
\newcommand{\csdc}{\mbox{$CS_{DC}$}}
\newcommand{\fl}{\mbox{\sc $FB$}}
\newcommand{\flS}{\mbox{\sc $FB_S$}}
\newcommand{\amongS}{\mbox{$AD_S$}}

\newcommand{\gsc}{\mbox{\sc Gsc}}
\newcommand{\gcc}{\mbox{\sc Gcc}}
\newcommand{\GCC}{\mbox{\sc Gcc}}
\newcommand{\AllDifferent}{\mbox{\sc AllDifferent}}

\newcommand{\nina}[1]{{#1}}

\newcommand{\SLIDE}{\mbox{\sc Slide}}
\newcommand{\SLIDINGSUM}{\mbox{\sc SlidingSum}}
\newcommand{\REGULAR}{\mbox{\sc Regular}}
\newcommand{\TABLE}{\mbox{\sc Table}}
\newcommand{\CIRCREGULAR}{\mbox{$\mbox{\sc Regular}_{\odot}$}}
\newcommand{\WRAPREGULAR}{\mbox{$\mbox{\sc Regular}^{*}$}}
\newcommand{\STRETCH}{\mbox{\sc Stretch}}
\newcommand{\INCSEQ}{\mbox{\sc IncreasingSeq}}
\newcommand{\INC}{\mbox{\sc Increasing}}
\newcommand{\lseX}{\mbox{\sc Lex}}
\newcommand{\NFA}{\mbox{\sc NFA}}
\newcommand{\DFA}{\mbox{\sc DFA}}

\newcommand{\PRECEDENCE}{\mbox{\sc Precedence}}
\newcommand{\lseXVAR}{\mbox{\sc LexLeader}}
\newcommand{\lseXGENSET}{\mbox{\sc SetGenLexLeader}}
\newcommand{\lseXSETVAR}{\mbox{\sc SetLexLeader}}
\newcommand{\lseXMSETVAR}{\mbox{\sc MSetLexLeader}}
\newcommand{\lseXSETVAL}{\mbox{\sc SetValLexLeader}}
\newcommand{\lseXVAL}{\mbox{\sc ValLexLeader}}
\newcommand{\lseXVALVAR}{\mbox{\sc GenLexLeader}}
\newcommand{\VALVARLEX}{\mbox{\sc ValVarLexLeader}}
\newcommand{\NVALUES}{\mbox{\sc NValues}}
\newcommand{\USES}{\mbox{\sc Uses}}
\newcommand{\COMMONG}{\mbox{\sc Common}}
\newcommand{\CARDPATH}{\mbox{\sc CardPath}}
\newcommand{\RANGE}{\mbox{\sc Range}}
\newcommand{\ROOTS}{\mbox{\sc Roots}}
\newcommand{\AMONG}{\mbox{\sc Among}}
\newcommand{\ATMOST}{\mbox{\sc AtMost}}
\newcommand{\ATLEAST}{\mbox{\sc AtLeast}}
\newcommand{\ATMOSTSEQ}{\mbox{\sc AtMostSeq}}
\newcommand{\ATLEASTSEQ}{\mbox{\sc AtLeastSeq}}
\newcommand{\AMONGSEQ}{\mbox{\sc AmongSeq}}
\newcommand{\SEQUENCE}{\mbox{\sc Sequence}}
\newcommand{\GENSEQUENCE}{\mbox{\sc Gen-Sequence}}
\newcommand{\SEQ}{\mbox{\sc Seq}}
\newcommand{\myelement}{\mbox{\sc Element}}
\newcommand{\LEX} {\mbox{\sc Lex}}
\newcommand{\REPEAT} {\mbox{\sc Repeat}}
\newcommand{\REPEATONE} {\mbox{\sc RepeatOne}}
\newcommand{\STRETCHREPEAT} {\mbox{\sc StretchRepeat}}
\newcommand{\STRETCHONEREPEAT} {\mbox{\sc StretchOneRepeat}}
\newcommand{\STRETCHONEREPEATONE} {\mbox{\sc StretchOneRepeatOne}}
\newcommand{\SETSIGLEX} {\mbox{\sc SetSigLex}}
\newcommand{\SETPREC} {\mbox{\sc SetPrecedence}}

\newcommand{\SOFTATMOSTSEC} {\mbox{\sc SoftAtMostSequence}}
\newcommand{\ATMOSTSEC} {\mbox{\sc AtMostSequence}}
\newcommand{\SOFTATMOST} {\mbox{\sc SoftAtMost}}
\newcommand{\SOFTSEQ} {\mbox{\sc SoftSequence}}
\newcommand{\SOFTAMONG} {\mbox{\sc SoftAmong}}
\newcommand{\SEQCYC}{\mbox{$\mbox{\sc CyclicSequence}$}}

\newcommand{\ATMOSTSEQCYC}{\mbox{$\mbox{\sc AtMostSeq}_{\odot}$}}
\newcommand{\ignore}[1]{}

{\makeatletter
 \gdef\xxxmark{%
   \expandafter\ifx\csname @mpargs\endcsname\relax 
     \expandafter\ifx\csname @captype\endcsname\relax 
       \marginpar{xxx}
     \else
       xxx 
     \fi
   \else
     xxx 
   \fi}
 \gdef\xxx{\@ifnextchar[\xxx@lab\xxx@nolab}
 \long\gdef\xxx@lab[#1]#2{{\bf [\xxxmark #2 ---{\sc #1}]}}
 \long\gdef\xxx@nolab#1{{\bf [\xxxmark #1]}}
  \long\gdef\xxx@lab[#1]#2{}\long\gdef\xxx@nolab#1{}%
}

\newcommand{\new}[1]{{#1}}
\newcommand{\myvec}[1]{\vec{#1}}
\pagestyle{plain}

\title{Flow-Based Propagators for the SEQUENCE \\ and Related 
Global Constraints\thanks{NICTA is funded by 
the Australian Government as represented by 
the Department of Broadband, Communications and the Digital Economy and 
the Australian Research Council. }}

\author{Michael Maher\inst{1} \and 
Nina Narodytska\inst{1} \and
Claude-Guy Quimper\inst{2} \and
Toby Walsh\inst{1}}
\institute{NICTA and UNSW, Sydney, Australia \and 
Ecole Polytechnique de Montreal, Montreal, Canada}

\maketitle

\begin{abstract}
We propose new filtering algorithms for the \SEQUENCE\ constraint
and some extensions of the \SEQUENCE\ constraint
based on
network flows. We enforce domain consistency
on the \SEQUENCE\ constraint  in $O(n^2)$
time down a branch of the search tree. This improves upon the best existing
domain consistency algorithm by a factor of $O(\log n)$.
\ignore{We also introduce a soft version of the \SEQUENCE\ constraint 
and propose an $O(n^2\log n \log \log u)$ time domain consistency algorithm
based on minimum cost network flows. }
The flows used in these algorithms
are 
derived from a 
linear program. Some of them differ from the flows used to propagate global
constraints like \gcc\ since the domains
of the variables are encoded as costs on 
the edges rather than capacities. 
Such flows
are efficient for maintaining bounds consistency over large domains
and may be useful for other global 
constraints. 
\end{abstract}

\section{Introduction}

Graph based algorithms play a very
important role in constraint programming,
especially within propagators for global constraints.
For example, Regin's propagator
for the \AllDifferent\ constraint
is based on a perfect matching algorithm \cite{Regin94},
whilst his propagator for the 
\GCC\ constraint is based on a 
network flow algorithm \cite{Regin96}. 
Both these graph algorithms are derived
from the bipartite value graph, in which
nodes represent variables and values, 
and edges represent domains. For example, the \GCC\ propagator
finds a flow in such a graph in which
each unit of flow represents
the assignment of a particular value to a variable. 
%
In this paper, we identify a new way to build
graph based propagators for global constraints:
we convert the global constraint into a linear
program and then convert this into a network flow.
{These encodings contain several novelties. For
example, variables domain bounds can be encoded as costs along the edges. 
}
We apply this approach to the \SEQUENCE\ family of constraints. 
Our results widen the class of 
global constraints which can be propagated
using flow-based algorithms. We conjecture
that these methods will be useful to 
propagate other global constraints.

\ignore{
After some background on constraints and the \SEQUENCE constraint,
in Section~\ref{sect:seqflow} we derive a network flow formulation that provides 
a domain consistency propagator for $\SEQUENCE$
of complexity $O(n^2)$ down a branch of a search tree.
We then define a soft sequence constraint and 
develop a $O(n^2 \log n \log \log u)$ propagator
based on minimum-cost flows.
In Section~\ref{sect:genseq} we discuss several methods that might be used to propagate
a generalized sequence constraint and provide a
$O(nW+n^2 \log n)$ domain consistency propagator
that can be improved to $O(n^2)$ in some circumstances.
We give an experimental evaluation of the propagators in Section \ref{sect:expt}.
}

\section{Background}

A constraint satisfaction problem (CSP) consists  of a set of
variables, each with a finite domain of values, and a set of
constraints specifying allowed combinations of values for
subsets of variables. We use capital letters for variables (e.g.
$X$, $Y$ and $S$), and lower case for values (e.g. $d$ and $d_i$).
A
solution is an assignment of values to the variables
satisfying the constraints.
Constraint solvers typically explore partial assignments enforcing
a local consistency property using either specialized or general
purpose propagation algorithms.
A \emph{support} for a constraint $C$ is a tuple 
that assigns a value to each variable from its domain which satisfies $C$.
A \emph{bounds support} is a tuple
that assigns a value to each variable which is between
the maximum and minimum in its domain which satisfies $C$.
A constraint is \emph{domain consistent} (\emph{DC}) iff for each
variable $X_i$, 
every value in the domain of $X_i$ belongs to a support.
A constraint is \emph{bounds consistent} (\emph{BC}) iff for each
variable $X_i$, 
there is a bounds support for the maximum and minimum value
in its domain. A CSP is \emph{DC}/\emph{BC} iff each constraint is \emph{DC}/\emph{BC}.
A constraint is \emph{monotone} iff there exists a total ordering
$\prec$ of the domain values such that for any two values $v$, $w$ if
$v \prec w$  then $v$ is substitutable for $w$ in any support for $C$.

 
 
We also give some background on flows. 
A {\it flow network} is a weighted directed graph $G=(V,E)$ where each edge $e$ has a capacity 
between non-negative integers $l(e)$ and $u(e)$,  and an integer cost $w(e)$.
A \textit{feasible flow} in a flow network between a source $(s)$ and a sink $(t)$, $(s,t)$-flow,  is a function $f: E\rightarrow  \ZZ^+$ that satisfies two conditions:
$f(e) \in [l(e), u(e)]$, $\forall e \in E$  and 
the \textit{flow conservation} law that ensures that
the amount of incoming flow should be equal to the amount of outgoing flow for all nodes except the source and the sink.  
The {\it value} of a $(s,t)$-flow is 
the amount of flow leaving the sink $s$. 
The {\it cost} of a flow $f$ is $w(f) = \sum_{e\in E} w(e)f(e)$.
A \textit{minimum cost flow}  is a feasible flow with the minimum cost. 
The Ford-Fulkerson algorithm can find a feasible flow in $O(\phi(f) |E|)$ time. 
If $w(e) \in  \ZZ$, $\forall e \in E$, then a minimum cost feasible flow can be found using the successive shortest path algorithm in $O(\phi(f) SPP)$ time, 
where $SPP$ is the complexity of finding a shortest path in the residual graph. 
Given a $(s,t)$-flow $f$ in $G(V,E)$, the {\it residual graph $G_f$} is the directed graph $(V,E_f)$, where
$E_f$ is
$$\{e\ \mbox{with cost}\ w(e)\ \mbox{and capacity}\ 0..(u(e)-f(e))~|~e=(u,v) \in E,  f(e)< u(e)\}  \bigcup$$     
$$\{e\ \mbox{with cost}\ -w(e)\ \mbox{and capacity}\ 0..(f(e)-l(e))~|~e=(u,v) \in E,  l(e)< f(e)\} $$     
There are other asymptotically faster but more complex algorithms for
finding either feasible or minimum-cost flows~\cite{AHUJA93}.

In our flow-based encodings, 
a consistency check will correspond to 
finding a feasible or minimum cost flow. 
To enforce \emph{DC}, we therefore need an algorithm
that, given a minimum cost flow of cost $w(f)$ and an edge $e$ checks if 
an extra unit flow can be pushed (or removed) through the edge $e$ and 
the cost of the resulting flow is less than or equal to a given threshold $T$. 
We use the residual graph to construct such an algorithm.
Suppose we need to check if an extra unit flow can be pushed through an edge $e = (u,v)$. 
Let $e'=(u,v)$ be the corresponding
arc in the residual graph.
If $w(e) = 0$, $\forall e \in E$, then it is sufficient to 
compute strongly connected components (SCC) in the residual graph.
An extra unit flow can be pushed through an edge $e$ iff both ends of the edge $e'$ are in the same strongly connected component. 
If $w(e) \in \ZZ$, $\forall e \in E$, the shortest path $p$ between $v$ and $u$
in the residual graph has to be computed. The minimal cost of pushing an extra unit 
flow through an edge $e$ equals $w(f) + w(p) + w(e)$. If $w(f) + w(p) + w(e) > T$,
then we cannot push an extra unit through $e$. Similarly, we can check if 
we can remove a unit flow through an edge. 

\section{The \SEQUENCE\ Constraint}

The $\SEQUENCE$ constraint was introduced
by Beldiceanu and Contejean~\cite{Beldiceanu94CHIP}.
It constrains the number of values
taken from a given set in any sequence of $k$ variables.
It is useful in staff rostering to specify, for example, that
every employee has at least 2 days off in any 7 day period.
Another application is sequencing cars along a production
line (prob001 in CSPLib).
It can specify, for example,
that at most 1 in 3 cars along the production line
has a sun-roof.
The $\SEQUENCE$ constraint can be defined in terms
of a conjunction of $\AMONG$ constraints. 
$\AMONG (l,u,[X_1,\ldots,X_k],v)$ holds
iff $l \leq |\{i|X_i\in v\}|\leq u$.
That is, between $l$ and $u$ of the $k$ variables take values in $v$.
The $\AMONG$ constraint can be encoded by channelling into 0/1 variables
using $Y_i\leftrightarrow (X_i\in v)$ and
$l \leq \sum_{i=1}^k Y_i \leq u$. Since the constraint graph
of this encoding is Berge-acyclic, this does not hinder propagation.
Consequently, 
we will simplify notation and consider $\AMONG$ (and \SEQUENCE)
on 0/1 variables and $v=\{1\}$.
If $l=0$, \AMONG\ is an \ATMOST\ constraint.
\ATMOST\ 
is {\em monotone} since,
given a support, we also have support for
any larger 
assignment~\cite{Bessiere07SLIDE}.
The $\SEQUENCE$ constraint is a conjunction
of overlapping $\AMONG$ constraints. More precisely, 
$\SEQUENCE(l,u,k,[X_1,\ldots,X_n],v)$
holds iff
for $1 \leq i \leq n-k+1$,
$\AMONG (l,u,[X_i,\ldots,X_{i+k-1}],v)$ holds.
A sequence like $X_i,\ldots,X_{i+k-1}$ is a \emph{window}.
It is easy to see that this decomposition
hinders propagation. 
If $l=0$, \SEQUENCE\ is an \ATMOSTSEQ\ constraint.
Decomposition in this case does not hinder propagation.
Enforcing \emph{DC} on the decomposition
of an \ATMOSTSEQ\ constraint
is equivalent to enforcing \emph{DC} on the $\ATMOSTSEQ$ constraint
~\cite{Bessiere07SLIDE}.


Several filtering algorithms exist
for $\SEQUENCE$ and related constraints. Regin and Puget
proposed a filtering algorithm for the Global Sequencing constraint (\gsc)
that combines a $\SEQUENCE$ and a global cardinality
constraint (\gcc)~\cite{Regin97GSC}. 
Beldiceanu and Carlsson suggested a greedy filtering algorithm for the
\CARDPATH\ constraint that can be used to propagate
the $\SEQUENCE$ constraint, but this 
may 
hinder propagation \cite{cardinality-path}.
Regin decomposed \gsc\ into a set of variable
disjoint \AMONG\ and \gcc\ constraints \cite{regin05}.
Again, this 
hinders propagation.
Bessiere {\it et al.}~\cite{Bessiere07SLIDE} encoded \SEQUENCE\ using
a \SLIDE\ constraint, and give a domain consistency propagator
that runs in $O(nd^{k-1})$ time.
van Hoeve {\it et al.}~\cite{Hoeve06}
proposed two filtering algorithms that establish
domain consistency. The first 
is based on an encoding into a $\REGULAR$ constraint
and runs in $O(n2^k)$ time,
whilst the second is based on 
cumulative sums and runs in $O(n^3)$ \nina {time down a branch of the search tree}. 
Finally, Brand {\it et al.}~\cite{bnqswcp07} studied
a number of different encodings 
of the $\SEQUENCE$
constraint. Their asymptotically 
fastest encoding is based on separation theory
and enforces domain consistency 
in $O(n^2 \log n)$ time down the whole branch of a search tree.
One of our contributions is to improve on this bound.

\section{Flow-based Propagator for the \SEQUENCE\ Constraint}   \label{sect:flowseq}

We will convert the \SEQUENCE\ constraint to a flow by means
of a linear program (LP). 
We shall use $\SEQUENCE(l,u,3,[X_1,\ldots,X_6],v)$
as a running example. 
We can formulate this constraint simply and directly as an integer linear program:
\begin{eqnarray*}
l \leq & X_1 + X_2 + X_3  & \leq u, \\
l \leq & X_2 + X_3 + X_4  & \leq u, \\
l \leq & X_3 + X_4 + X_5  & \leq u, \\
l \leq & X_4 + X_5 + X_6  & \leq u 
\end{eqnarray*}
where $X_i \in \{0, 1\}$.
By introducing surplus/slack variables,
$Y_i$ and $Z_i$, we convert this to a set of equalities:
\begin{eqnarray*}
X_1 + X_2 + X_3 - Y_1 = l, & \ \ \ \ & X_1 + X_2 + X_3 + Z_1 = u, \\
X_2 + X_3 + X_4 - Y_2 = l, & & X_2 + X_3 + X_4 + Z_2 = u, \\
X_3 + X_4 + X_5 - Y_3 = l, & & X_3 + X_4 + X_5 + Z_3 = u, \\
X_4 + X_5 + X_6 - Y_4 = l, & & X_4 + X_5 + X_6 + Z_4 = u 
\end{eqnarray*}
where  $Y_i, Z_i \geq 0$. 
In matrix form, this is:
$$
\left(
\begin{smallmatrix}
  1 & 1 & 1 & 0 & 0 & 0 &  -1 & 0&  0& 0&  0& 0&  0& 0\\
  1 & 1 & 1 & 0 & 0 & 0 &  0  & 1&  0& 0&  0& 0&  0& 0\\
  0 & 1 & 1 & 1 & 0 & 0 &  0  & 0& -1& 0&  0& 0&  0& 0\\
  0 & 1 & 1 & 1 & 0 & 0 &  0  & 0&  0& 1&  0& 0&  0& 0\\
  0 & 0 & 1 & 1 & 1 & 0 &  0  & 0&  0& 0& -1& 0&  0& 0\\
  0 & 0 & 1 & 1 & 1 & 0 &  0  & 0&  0& 0&  0& 1&  0& 0\\
  0 & 0 & 0 & 1 & 1 & 1 &  0  & 0&  0& 0&  0& 0& -1& 0\\
  0 & 0 & 0 & 1 & 1 & 1 &  0  & 0&  0& 0&  0& 0&  0& 1 
\end{smallmatrix}
\right)
\left(
\begin{smallmatrix}
X_1 \\ 
\vdots \\
X_6 \\ 
Y_1 \\
Z_1 \\
\vdots \\
Y_4 \\
Z_4 
\end{smallmatrix}
\right)
=
\left(
\begin{smallmatrix}
l \\
u \\
l \\ 
u \\ 
l \\
u \\ 
l \\
u 
\end{smallmatrix}
\right)
$$
This matrix has the \emph{consecutive ones} property for columns:
each column has a block of consecutive 1's or $-1$'s
and the remaining elements are 0's.
Consequently, we can apply the method of Veinott and Wagner \cite{VWa62}
(also described in Application 9.6 of ~\cite{AHUJA93}) to simplify the problem.
We create 
a zero last row
and subtract the $i$th row from $i+1$th row for 
$i=1$ to $2n$. These operations
do not change the set of solutions.
This gives:
$$
\left(
\begin{smallmatrix}
  1 & 1 & 1 & 0 & 0 & 0 &  -1 & 0&  0& 0&  0& 0&  0& 0\\
  0 & 0 & 0 & 0 & 0 & 0 &  1  & 1&  0& 0&  0& 0&  0& 0\\
  -1 & 0 & 0 & 1 & 0 & 0 &  0  & -1& -1& 0&  0& 0&  0& 0\\
  0 & 0 & 0 & 0 & 0 & 0 &  0  & 0&  1& 1&  0& 0&  0& 0\\
  0 & -1 & 0 & 0 & 1 & 0 &  0  & 0&  0& -1& -1& 0&  0& 0\\
  0 & 0 & 0 & 0 & 0 & 0 &  0  & 0&  0& 0&  1& 1&  0& 0\\
  0 & 0 & -1 & 0 & 0 & 1 &  0  & 0&  0& 0&  0& -1& -1& 0\\
  0 & 0 & 0 & 0 & 0 & 0 &  0  & 0&  0& 0&  0& 0&  1& 1  \\
  0 & 0 & 0 & -1 & -1 & -1 &  0  & 0&  0& 0&  0& 0&  0& -1  \\
\end{smallmatrix}
\right)
\left(
\begin{smallmatrix}
X_1 \\ 
\vdots \\
X_6 \\ 
Y_1 \\
Z_1 \\
\vdots \\
Y_4 \\
Z_4 
\end{smallmatrix}
\right)
=
\left(
\begin{smallmatrix}
l \\  u-l \\   l-u  \\  u-l \\  l-u \\ u-l \\ l-u \\ u-l \\ -u
\end{smallmatrix}
\right)
$$

This matrix has a single $1$ and $-1$ in each column.
Hence, 
it describes a network flow problem~\cite{AHUJA93} on a graph
$G = (V,E)$ (that is, it is a network matrix).
Each row in the matrix 
corresponds to a node in $V$ and each
column 
corresponds to an edge in $E$. 
Down each column, there is a single
row $i$ equal to 1 and a single row $j$ equal to -1 corresponding
to an edge $(i, j) \in E$ in the graph.
We include a source node $s$ and a sink node $t$ in $V$. Let
$b$ be the vector on the right hand side of the equation. If $b_i$ is
positive, then there is an edge $(s, i) \in E$ that carries exactly
$b_i$ amount of flow. If $b_i$ is negative, there is an edge $(i, t)
\in E$ that caries exactly $|b_i|$ amount of flow.
The bounds on the variables, which are not expressed in the matrix,
are represented as bounds on the capacity of the corresponding edges.

\begin{figure}[htb] \centering
  \vspace{-3.2cm}
    \includegraphics[width=0.7\textwidth]{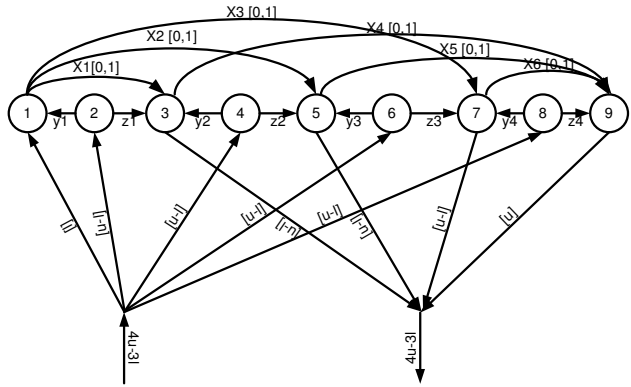}\\
    \vspace{-2.8cm}
    \caption{\label{f:f1}
    A flow graph for $\SEQUENCE(l,u,3,[X_1,\ldots,X_6],v)$}
\end{figure}

The graph for the set of equations in the example is given in
Figure~\ref{f:f1}.  A flow of value $4u-3l$ in the graph corresponds
to a solution.  If a feasible flow sends a unit flow through the edge
labeled with $X_i$ then $X_i = 1$ in the solution; otherwise $X_i = 0$.
Each even numbered vertex $2i$ represents a window.
The way the incoming flow is shared between $y_{j}$ and $z_{j}$ reflects
how many variables $X_i$ in the $j$'th window are equal to 1.
Odd numbered vertices represent transitions from one window to the next
(except for the first and last vertices, which represent transitions between a window and nothing).
An incoming $X$ edge represents the variable omitted in the transition to the next window,
while an outgoing $X$ edge represents the added variable.

\begin{theorem}
\label{t:seq_flow}
For any constraint $\SEQUENCE(l,u,k,[X_1,\ldots,X_n],v)$,
there is an equivalent network flow graph  $G = (V,E)$ with
$5n -4k + 5$ edges, 
$2n-2k + 3 +2$ vertices,
a maximum edge capacity of $u$, 
and an amount of flow to send equal to $f = (n-k)(u-l)+u$.
There is a one-to-one correspondence between solutions of the constraint
and feasible flows in the network.
\end{theorem}
\sloppy{
The time complexity of finding a maximum flow 
of  value $f$ is $O(|E|f)$ using the Ford-Fulkerson algorithm \cite{Cormen96}. 
Faster algorithms exist for this problem. 
For example,
Goldberg and Rao's algorithm finds a maximum flow in
$O(min(|V|^{2/3},|E|^{1/2})|E|\log (|V|^2/|E| +2)\log C)$ time 
where $C$ is the maximum capacity upper bound for an edge
\cite{Goldberg98}. 
In our case, this gives $O(n^{3/2}\log n \log u)$ time complexity.
}
We follow R\'egin~\cite{Regin94,Regin96} in the building of an incremental filtering algorithm
from the network flow formulation.
A feasible flow in the graph gives us a support
for one value in each variable domain. Suppose $X_k = v$ is in the
solution that corresponds to the feasible flow where $v$ is either zero or one. 
To obtain a support for $X_k=1-v$, we find the SCC of the residual graph and
check if both ends of the edge labeled with $X_k$ are in the same
strongly connected component. If so,  $X_k = 1-v$ has a support; otherwise
$1-v$ can be removed from the domain of $X_k$.
\nina {Strongly connected components can be found in linear time, because 
the number of nodes and edges in the flow network for the $\SEQUENCE$ constraint is linear in $n$ by Theorem~\ref{t:seq_flow}.}
The total time complexity for initially enforcing \emph{DC} is 
$O(n((n-k)(u-l) + u))$ if we use the Ford-Fulkerson algorithm or 
$O(n^{3/2}\log n \log u)$ if we use Goldberg and Rao's algorithm. 

Still following R\'egin~\cite{Regin94,Regin96}, one can make the algorithm incremental. 
Suppose during search $X_i$
is fixed to value $v$. If the last computed flow was a support for
$X_i = v$, then there is no need to recompute the flow. We simply need
to recompute the SCC in the new residual graph and enforce \emph{DC} in
$O(n)$ time. If the last computed flow is not a support for $X_i = v$,
we can find a cycle in the residual graph 
containing the edge associated to $X_i$ in $O(n)$ time. By pushing a unit of flow over
this cycle, we obtain a flow that is a support for $X_i =
v$. Enforcing \emph{DC} can be done in $O(n)$ after computing the SCC.
Consequently, there is an incremental cost of $O(n)$ when a variable is fixed,
and the cost of enforcing \emph{DC} down a branch of the search tree is $O(n^2)$.

\section{Soft \SEQUENCE\  Constraint}

Soft forms
of the \SEQUENCE\ constraint may be useful in practice. 
The ROADEF 2005 challenge \cite{roadef}, which
was proposed and sponsored by Renault,
puts forward a violation measure for
the \SEQUENCE\ constraint which 
takes into account by how much each \AMONG\
constraint is violated. 
We therefore consider the soft global constraint,
$\SOFTSEQ(l,u,k,T,[X_1,\ldots, X_n],v)$. This
holds iff:
\begin{eqnarray}
T &\geq &\sum_{i=1}^{n-k+1}
\mymax(l-\sum_{j=0}^{k-1} (X_{i+j} \in v),
\sum_{j=0}^{k-1} (X_{i+j}\in v)-u,0) \label{eqn::defSoftSeq}
\end{eqnarray}
As before, we 
can simplify notation and consider \SOFTSEQ\
on 0/1 variables and $v=\{1\}$.

We again convert to a flow problem by means of a
linear program, but this time with an objective function.  Consider
$\SOFTSEQ(l,u,3,T,[X_1,\ldots, X_6],v)$.  We introduce variables,
$Q_i$ and $P_i$ to represent the penalties that may arise from
violating lower and upper bounds respectively.  We can then express
this $\SOFTSEQ$ constraint as follows. The objective function gives a
lower bound on $T$. 
\xxx[CG]{What does ``where we minimize the lower
  bound on $T$'' mean in this context?}  \xxx[MM]{I added this because
  we need to link T to the formulation below.  Any improvement
  welcome} \xxx[CG]{How about this version?}
\begin{eqnarray*}
{\rm Minimize} \sum_{i=1}^{4}(P_i+Q_i) & & {\rm subject}  \ {\rm to:}\\
X_1 + X_2 + X_3 - Y_1 + Q_1 = l, & \ \ \ \ & X_1 + X_2 + X_3 + Z_1 - P_1 = u, \\
X_2 + X_3 + X_4 - Y_2 + Q_2 = l, & & X_2 + X_3 + X_4 + Z_2 - P_2 = u, \\
X_3 + X_4 + X_5 - Y_3 + Q_3 = l, & & X_3 + X_4 + X_5 + Z_3 - P_3 = u, \\
X_4 + X_5 + X_6 - Y_4 + Q_4 = l, & & X_4 + X_5 + X_6 + Z_3 - P_4 = u 
\end{eqnarray*}
\noindent
where $Y_i$, $Z_i$, $P_i$ and $Q_i$ are non-negative.
In matrix form, this is:

\vspace{-3em}
$$
\begin{tabular}{c} 
{\rm Minimize} \mbox{$\sum_{i=1}^{4}(P_i+Q_i)$} \ \ \ {\rm subject}  \ {\rm to:}\\
\ \\
\mbox{$
\left(
\begin{smallmatrix}
  1 & 1 & 1 & 0 & 0 & 0 &  -1 & 0&  0& 0&  0& 0&  0& 0 &  1 & 0&  0& 0&  0& 0&  
0& 0\\
  1 & 1 & 1 & 0 & 0 & 0 &  0  & 1&  0& 0&  0& 0&  0& 0 &  0  & -1&  0& 0&  0& 0&
  0& 0\\
  0 & 1 & 1 & 1 & 0 & 0 &  0  & 0& -1& 0&  0& 0&  0& 0 &  0  & 0& 1& 0&  0& 0&  
0& 0\\
  0 & 1 & 1 & 1 & 0 & 0 &  0  & 0&  0& 1&  0& 0&  0& 0 &  0  & 0&  0& -1&  0& 0&
  0& 0\\
  0 & 0 & 1 & 1 & 1 & 0 &  0  & 0&  0& 0& -1& 0&  0& 0 &  0  & 0&  0& 0&  1& 0& 
 0& 0\\
  0 & 0 & 1 & 1 & 1 & 0 &  0  & 0&  0& 0&  0& 1&  0& 0 &  0  & 0&  0& 0&  0& -1&
  0& 0\\
  0 & 0 & 0 & 1 & 1 & 1 &  0  & 0&  0& 0&  0& 0& -1& 0 &  0  & 0&  0& 0&  0& 0& 
1& 0\\
  0 & 0 & 0 & 1 & 1 & 1 &  0  & 0&  0& 0&  0& 0&  0& 1 &  0  & 0&  0& 0&  0& 0& 
 0& -1 
\end{smallmatrix}
\right)
$}
\end{tabular}
\begin{tabular}{c} 
\ \\
\ \\
\mbox{$
\left(
\begin{smallmatrix}
X_1 \\ 
\vdots \\
X_6 \\ 
Y_1 \\
Z_1 \\
\vdots \\
Y_4 \\
Z_4 \\
Q_1 \\
P_1 \\
\vdots \\
Q_4 \\
P_4 
\end{smallmatrix}
\right)
=
\left(
\begin{smallmatrix}
l \\
u \\
l \\ 
u \\ 
l \\
u \\ 
l \\
u 
\end{smallmatrix}
\right)
$}
\end{tabular}
$$
If we transform the matrix as before, we get a 
minimum cost network flow
problem:

\vspace{-3em}
$$
\begin{tabular}{c} 
{\rm Minimize} \mbox{$\sum_{i=1}^{4}(P_i+Q_i)$} \ \ \ {\rm subject}  \ {\rm to:}\\
\ \\
\mbox{$
\left(
\begin{smallmatrix}
  1 & 1 & 1 & 0 & 0 & 0 &  -1 & 0&  0& 0&  0& 0&  0& 0 &  1 & 0&  0& 0&  0& 0&  
0& 0\\
  0 & 0 & 0 & 0 & 0 & 0 &  1  & 1&  0& 0&  0& 0&  0& 0 &  -1  & -1&  0& 0&  0& 0
&  0& 0\\
  -1 & 0 & 0 & 1 & 0 & 0 &  0  & -1& -1& 0&  0& 0&  0& 0 &  0  & 1& 1& 0&  0& 0&
  0& 0\\
  0 & 0 & 0 & 0 & 0 & 0 &  0  & 0&  1& 1&  0& 0&  0& 0 &  0  & 0&  -1& -1&  0& 0
&  0& 0\\
  0 & -1 & 0 & 0 & 1 & 0 &  0  & 0&  0& -1& -1& 0&  0& 0&  0  & 0&  0& 1& 1& 0& 
 0& 0\\
  0 & 0 & 0 & 0 & 0 & 0 &  0  & 0&  0& 0&  1& 1&  0& 0 &  0  & 0&  0& 0&  -1& -1
&  0& 0\\
  0 & 0 & -1 & 0 & 0 & 1 &  0  & 0&  0& 0&  0& -1& -1& 0 &  0  & 0&  0& 0&  0& 1
& 1& 0\\
  0 & 0 & 0 & 0 & 0 & 0 &  0  & 0&  0& 0&  0& 0&  1& 1 &  0  & 0&  0& 0&  0& 0& 
 -1& -1 \\
  0 & 0 & 0 & -1 & -1 & -1 &  0  & 0&  0& 0&  0& 0&  0& -1 &  0  & 0&  0& 0&  0&
 0&  0& 1  
\end{smallmatrix}
\right)
$}
\end{tabular}
\begin{tabular}{c} 
\ \\
\ \\
\mbox{$
\left(
\begin{smallmatrix}
X_1 \\ 
\vdots \\
X_6 \\ 
Y_1 \\
Z_1 \\
\vdots \\
Y_4 \\
Z_4 \\
Q_1 \\
P_1 \\
\vdots \\
Q_4 \\
P_4 
\end{smallmatrix}
\right)
=
\left(
\begin{smallmatrix}
l \\  u-l \\   l-u  \\  u-l \\  l-u \\ u-l \\ l-u \\ u-l \\ -u
\end{smallmatrix}
\right)
$}
\end{tabular}
$$

\begin{figure}[htb] \centering
  \vspace{-3.2cm}
    \includegraphics[width=0.7\textwidth]{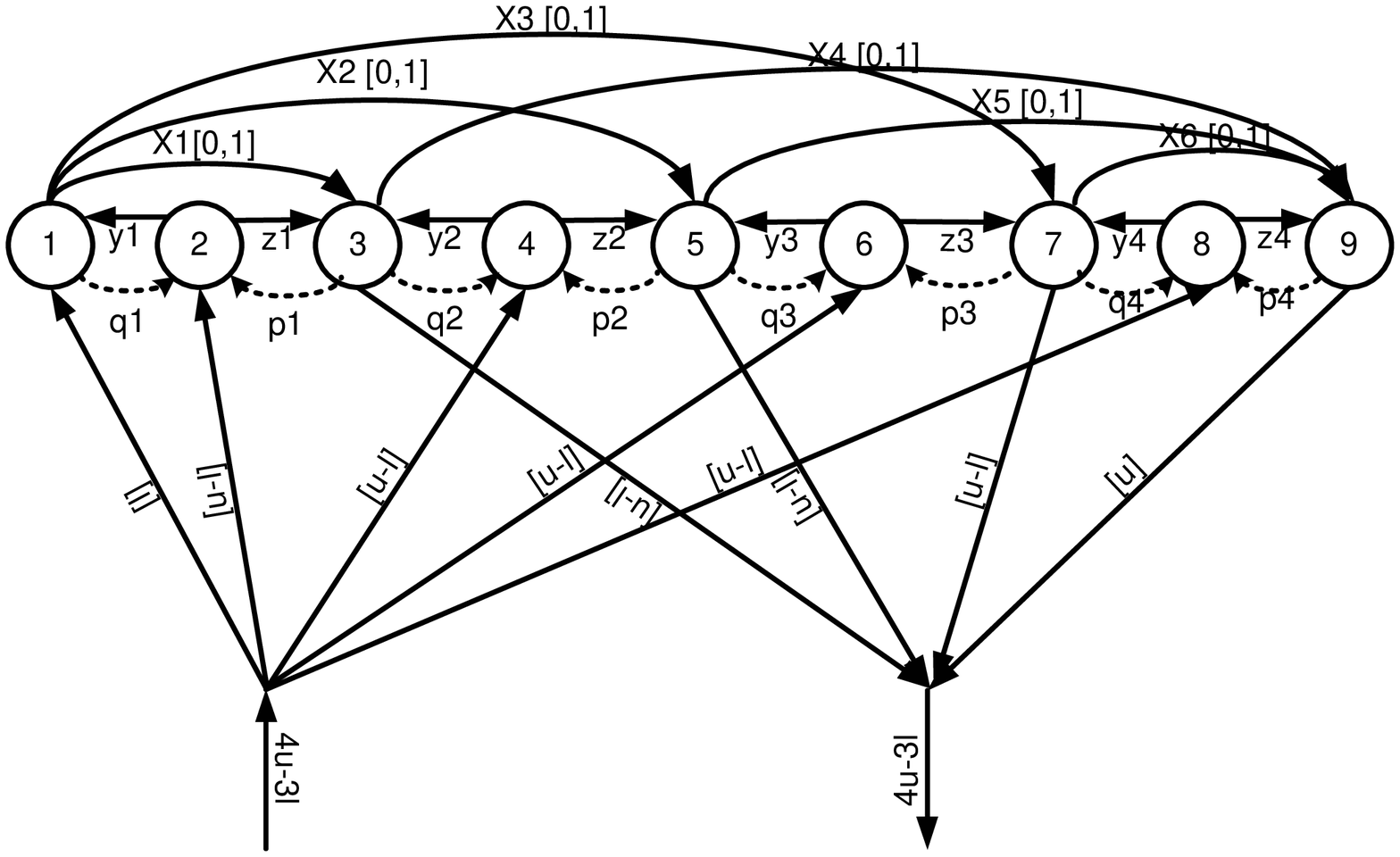}\\
    \vspace{-2.8cm}
    \caption{\label{f:f2}
    A flow graph for $\SOFTSEQ(l,u,3,T,[X_1,\ldots,X_6])$}
\end{figure}

The flow graph $G = (V,E)$ for this
system is presented  in Figure~\ref{f:f2}.
Dashed edges have cost $1$, 
while other edges have cost $0$.
The minimal cost flow in the graph corresponds to a minimal cost
solution to the system of equations

\begin{theorem}
\label{t:soft_seq}
For any constraint $\SOFTSEQ(l,u,k,T,[X_1,\ldots,X_n],v)$,
there is an equivalent network flow graph.  
There is a one-to-one correspondence between solutions of the constraint
and feasible flows 
of cost less than or equal to $max(dom(T))$. 
\end{theorem}

\nina{ 
Using Theorem~\ref{t:soft_seq}, we construct a \emph{DC} filtering algorithm for the $\SOFTSEQ$ constraint.
The  $\SOFTSEQ$ constraint is \emph{DC} iff the following conditions hold:
\begin{itemize}
	\item value $1$ belongs to $dom(X_i)$, $i=1,\ldots,n$ iff
there exists a feasible flow  of cost at most $\max(\dom(T))$ that sends a unit flow through the edge labeled with $X_i$.
	\item value $0$ belongs to $dom(X_i)$, $i=1,\ldots,n$ iff
there exists a feasible flow  of cost at most $\max(\dom(T))$ that does not send any flow through the edge labeled with $X_i$.
	\item there exists a feasible flow of cost at most $\min(\dom(T))$.
\end{itemize}
}

The minimal cost flow 
can be found in 
$O(|V||E| \log \log U \log |V|C) = O (n^2\log n \log \log u)$
time~\cite{AHUJA93}. 
Consider the edge $(u, v)$ in the residual graph associated to
variable $X_i$ and let $k_{(u,v)}$ be its residual cost. If the flow
corresponds to an assignment with $X_i = 0$, pushing a unit of flow
on $(u, v)$ results in a solution with $X_i = 1$. Symmetrically, if
the flow corresponds to an assignment with $X_i = 1$, pushing a unit
of flow on $(u, v)$ results in a solution with $X_i = 0$. If the
shortest path in the residual graph between $v$ and $u$ is
$k_{(v,u)}$, then the shortest cycle that contains $(u, v)$ has length
$k_{(u,v)} + k_{(v,u)}$. Pushing a unit of flow through this cycle
results in a flow of cost $c + k_{(u,v)} + k_{(v,u)}$ which is the
minimum-cost flow that contains the edge $(u, v)$. If $c + k_{(u,v)} +
k_{(v,u)} > \max(\dom(T))$, then no flows containing the edge $(u, v)$
exist with a cost smaller or equal to $\max(\dom(T))$. The variable
$X_i$ must therefore be fixed to the value taken in the current flow.
Following Equation~\ref{eqn::defSoftSeq}, the cost of the variable $T$
must be no smaller than the cost of the solution. To enforce \emph{BC} 
on the cost variable, we increase the lower bound of
$\dom(T)$ to the cost of the minimum flow in the graph $G$.

To enforce \emph{DC} on the $X$ variables efficiently 
we can use an
all pairs shortest path algorithm 
on the residual graph~\cite{Regin99b}.
This takes
$O(n^2\log n)$ time using Johnson's algorithm~\cite{Cormen96}. This
gives an $O(n^2 \log n \log \log u)$ time complexity to enforce \emph{DC}
on $\SOFTSEQ$.
The penalty variables used for $\SOFTSEQ$ arise directly out of the
problem description and occur naturally in the LP formulation.  We
could also view them as arising through the methodology of \cite{HPR06},
where edges with costs are added to the network graph for the hard
constraint to represent the softened constraint. 

\ignore{
\subsection{Soft \ATMOSTSEQ\ Constraint}

In many cases, we have only
upper bounds and not lower bounds on the frequency
of the occurrence of values (i.e. $l=0$). 
For instance, this is the case in car sequencing problems.
This can be used to simplify propagation.
For example, there is a simple
propagator to enforce \emph{DC} on the 
soft \ATMOSTSEQ\ constraint in just 
$O(n^2 k)$ time down a branch of the search tree.
Consider the assignment which assigns each $X_i$ the smallest value in
its domain. Due to monotonicity of the $\ATMOSTSEQ$ constraint any
other solution $X_i'$ will be greater or equal to this minimal
assignment: $X_i \leq X_i'$, $i=1,\ldots,n$.  The violation measure is
a monotonically non-decreasing function of the $X_i$.  Consequently,
the violation cost for any other solution is greater or equal to the
violation cost of this minimal assignment.  Hence, if the violation
cost for the minimal assignment is greater than the upper bound on
the cost variable then the constraint is inconsistent.  To enforce
\emph{DC} on soft $\ATMOSTSEQ$, we can use the failed literal test. If a
value is pruned from the domain of $X_i$, then it takes $O(k)$ time to
update the  cost value of the minimal assignment and $O(nk)$ 
time to perform the failed literal test for $n$ Boolean variables. 
Hence, the total time complexity is $O(n^2k)$ down a branch of the search tree.
}

\ignore{
\xxx[MM]{The text below is on borrowed time}

Another possibility is to use
the decomposition of the soft $\ATMOSTSEQ$ constraint
into soft $\ATMOST$ constraints and a sum of their
individual costs. Whilst this decomposition hinders propagation,
it is correct for disentailment. 
\begin{lemma}
		\label{l:l3}
		The decomposition of the soft $\ATMOSTSEQ([X_1,\ldots,X_n],u, k, T)$ constraint into
		individual soft $\ATMOST([X_i,\ldots,X_{i+k-1}],u, k, T_i)$, $i=1,\ldots,n-k+1$ constraints
		and the linear constraint $\sum_{i=1}^{n-k+1}{T_i} \leq T$ is correct for disentailment.
		\end{lemma}		 
		\proof 
		 Let soft $\ATMOST([X_i,\ldots,X_{i+k-1}],u,T_i)$, $i=1,\ldots,n-k+1$
     and $\sum_{i=1}^{n-p+1}{T_i} \leq T$ be bounds consistent. We need show how to construct 
     a solution of the soft $\ATMOSTSEQ$ constraint.
       
     Consider the minimal assignment $X_j = min (D(X_j))$, $j=1,\ldots,n$. 
     This assignment satisfies each soft $\ATMOST$ constraint and the cost of 
     it equals  $\sum_{i=1}^{n-k+1}{lb(T_i)}$ that is less than
     or equal to $T$. Hence, the minimal assignment satisfies the 
     soft $\ATMOSTSEQ$ constraint.       
  \qed 

Consequently, \emph{DC} on soft $\ATMOSTSEQ$ can be achieved
by enforcing singleton bounds consistency on the decomposition.

\begin{proposition}
Consider the decomposition of the soft $\ATMOSTSEQ([X_1,\ldots,X_n],u, k, T)$ constraint into
individual soft $\ATMOST([X_i,\ldots,X_{i+k-1}],u, k, T_i)$, $i=1,\ldots,n-k+1$ constraints
and the linear constraint $\sum_{i=1}^{n-k+1}{T_i} \leq T$.
Domain consistency can be achieved
by enforcing singleton bounds consistency on this decomposition.
\end{proposition}		 
}

\section{Generalized \SEQUENCE\ Constraint}   \label{sect:genseq}

To model real world problems, we may want to
have different size or positioned windows. For example, 
the window size in a rostering problem may
depend on whether it includes a weekend or not. 
An extension of the \SEQUENCE\ constraint
proposed in ~\cite{Hoeve06}
is that each \AMONG\ constraint can have different
parameters (start position, $l$, $u$, and $k$). 
More precisely, 
$\GENSEQUENCE(\myvec{p_1}, 
\ldots,
\myvec{p_m},
[X_1,X_2,\ldots,X_n],v)$
holds
iff 
$\AMONG(l_i,u_i,k_i,[X_{s_i},\ldots,X_{s_i+k_i-1}],v)$
for $1 \leq i \leq m$
where $\myvec{p_i} = \left\langle l_i,u_i,k_i,s_i \right\rangle$. 
Whilst the methods in Section~\ref{sect:flowseq} easily extend
to allow different bounds $l$ and $u$ for each window, dealing with
different windows is more difficult. In general, 
the matrix now does not have the consecutive ones property. 
It may be possible to re-order the windows
to achieve the consecutive ones property. 
If such a re-ordering exists, 
it can be found and performed in $O(m + n + r)$ time, 
where $r$ is the number of non-zero entries in the matrix \cite{BL76}.
Even when re-ordering cannot achieve the consecutive ones property
there may, nevertheless, be an equivalent network matrix.
Bixby and Cunningham \cite{Bixby80} give a procedure to find
an equivalent network matrix, when it exists, in $O(mr)$ time.
Another procedure is given in \cite{Schrijver87}.
In these cases, the method
in Section~\ref{sect:flowseq} can be applied to
propagate the $\GENSEQUENCE$ constraint in $O(n^2)$ time 
down the branch of a search tree.

Not all $\GENSEQUENCE$ constraints can be
expressed as network flows. 
Consider the 
$\GENSEQUENCE$ constraint
with $n=5$, identical upper and lower
bounds ($l$ and $u$), and 4 windows: [1,5], [2,4], [3,5], 
and [1,3]. 
We can express it as
an integer linear program:
\begin{eqnarray}
\left(
\begin{smallmatrix}
  1 & 1 & 1 & 1 & 1 \\
  -1 & -1 & -1 & -1 & -1 \\
  0 & 1 & 1 & 1 & 0 \\
  0 & -1 & -1 & -1 & 0 \\
  0 & 0 & 1 & 1 & 1 \\
  0 & 0 & -1 & -1 & -1 \\
  1 & 1 & 1 & 0 & 0 \\
  -1 & -1 & -1 & 0 & 0
\end{smallmatrix}
\right)
\left(
\begin{smallmatrix}
X_1 \\ 
X_2 \\
X_3 \\ 
X_4 \\
X_5 
\end{smallmatrix}
\right)
\geq
\left(
\begin{smallmatrix}
l \\
-u \\
l \\ 
-u \\ 
l \\
-u \\ 
l \\
-u 
\end{smallmatrix}
\right) \label{eqn::exampleGen}
\end{eqnarray}
Applying the test described in Section 20.1
of~\cite{Schrijver87} to Example~\ref{eqn::exampleGen}, we find that the matrix of this problem is not equivalent to any network matrix.


However, all $\GENSEQUENCE$ constraint matrices satisfy the weaker property of
total unimodularity.
A matrix is \emph{totally unimodular} iff
every square non-singular submatrix has a determinant
of $+1$ or $-1$. 
The advantage of this property is that any totally unimodular
system of inequalities with integral constants is solvable in $\mathbb{Z}$
iff it is solvable in $\mathbb{R}$.

\begin{theorem}
  The matrix of the inequalities associated with
  \GENSEQUENCE\ constraint is totally unimodular.
\end{theorem}
\xxx[CG]{I feel we need a proof here. However, the proof written in the tex file does not seem appropriate in this context.}
\ignore{
\begin{proof}
 The transpose of the constraint matrix has the 
  the consecutive ones property. 
 The transpose is therefore totally unimodular. A matrix is
  totally unimodular if and only if its transpose is totally
  unimodular. Therefore, the original matrix is totally
unimodular. 
\end{proof}
}

In practice, only integral values for the bounds $l_i$ and $u_i$ are used.
Thus the consistency of a \GENSEQUENCE\ constraint
can be determined via \nina{interior point algorithms for LP} in
$O(n^{3.5} \log u)$ time. 
Using the failed literal test,
we can enforce \emph{DC} at a cost of $O(n^{5.5} \log u)$ down the branch of a search tree
for any \GENSEQUENCE\ constraint.
This is too expensive to be practical. 
We can, instead, exploit the fact that the matrix for each \GENSEQUENCE\ constraint
has the consecutive ones property \emph{for rows}
(before the introduction of slack/surplus variables).
Corresponding to the row transformation
for matrices with consecutive ones for columns
is a change-of-variables transformation into variable $S_j = \sum_{i=1}^j X_i$
for matrices with consecutive ones for rows.
This gives the dual of a network matrix.
This is the basis of an encoding of \SEQUENCE\ in
\cite{bnqswcp07} (denoted there $CD$).
Consequently that encoding extends to \GENSEQUENCE.
Adapting the analysis in \cite{bnqswcp07} to  \GENSEQUENCE,
we can enforce \emph{DC} in
$O(nm + n^2\log n)$ time down the branch of a search tree.

In summary, for a compilation cost of $O(mr)$, we can enforce \emph{DC} on a
\GENSEQUENCE\ constraint in $O(n^2)$ down the branch of a search tree, 
when it has a flow representation,
and in $O(nm + n^2\log n)$ when it does not.

\section{A \SLIDINGSUM\ Constraint}

The \SLIDINGSUM\ constraint \cite{Beldiceanu06CAT} is a generalization of
the \SEQUENCE\ constraint from Boolean to integer variables, 
which we extend to allow arbitrary windows.
$\SLIDINGSUM$ $([X_1, \ldots, X_n],[\myvec{p_1}, \ldots, \myvec{p_m}])$
holds iff $l_i \leq \sum_{j=s_i}^{s_i + k_i -1} X_i \leq u_i$ holds
where $\myvec{p_i} = \left\langle l_i,u_i,k_i,s_i \right\rangle$ is,
as with the generalized \SEQUENCE, a window. The constraint can be
expressed as a linear program $\mathcal{P}$ called the \emph{primal}
where $W$ is a matrix encoding the inequalities. Since the constraint
represents a satisfaction problem, we minimize the constant 0. The
\emph{dual} $\mathcal{D}$ is however an optimization problem.

\begin{align}
  \left.
    \begin{aligned}
      \min\; 0 \\
      \begin{bmatrix}
        W \\
        -W \\
        I \\
        -I
      \end{bmatrix}
      X
      & \geq
      \begin{bmatrix}
        l \\
        -u \\
        a \\
        -b
      \end{bmatrix}
    \end{aligned}
  \hspace{5mm} \right\} \mathcal{P} & \hspace{5mm} &
  \left.
    \begin{aligned}
      \min \begin{bmatrix} -l & u & -a & b \end{bmatrix} Y \\
      \begin{bmatrix}
        W^T & - W^T & I & - I
      \end{bmatrix}
      Y
      & =
      0 \\
      Y & \geq 0
    \end{aligned}
  \hspace{5mm} \right\} \mathcal{D}
\end{align}

Von Neumann's Strong Duality Theorem states that if the primal and the
dual problems are feasible, then they have the same objective
value. Moreover, if the primal is unsatisfiable, the dual is
unbounded. The \SLIDINGSUM\ constraint is thus satisfiable if the
objective function of the dual problem is zero. It is unsatisfiable if
it tends to negative infinity.

Note that the matrix $W^T$ has the consecutive ones property on the
columns. The dual problem can thus be converted to a network flow
using the same transformation as with the \SEQUENCE\ constraint.
Consider the dual LP of our running example:
$$
\begin{tabular}{c} 
{\rm Minimize} \mbox{$-\sum_{i=1}^{4} l_i Y_{i}+ \sum_{i=1}^4 u_i Y_{4 +i} - \sum_{i=1}^5 a_i Y_{8+i} + \sum_{i=1}^5 b_i Y_{13+i}$} \ \ \ {\rm subject}  \ {\rm to:}\\
\ \\
\mbox{$
\left(
\begin{smallmatrix}
1 & 0 & 0 & 1 &  -1 & 0  & 0  &-1 & 1 & 0 & 0 & 0 & 0 & -1 & 0 & 0 & 0 & 0\\
1 & 1 & 0 & 1 &  -1 & -1 & 0  &-1 & 0 & 1 & 0 & 0 & 0 & 0 & -1 & 0 & 0 & 0\\
1 & 1 & 1 & 1 &  -1 & -1 & -1 &-1 & 0 & 0 & 1 & 0 & 0 & 0 & 0 & -1 & 0 & 0\\
1 & 1 & 1 & 0 &  -1 & -1 & -1 &0  & 0 & 0 & 0 & 1 & 0 & 0 & 0 & 0 & -1 & 0\\
1 & 0 & 1 & 0 &  -1 & 0  & -1 &0  & 0 & 0 & 0 & 0 & 1 & 0 & 0 & 0 & 0 & -1\\
\end{smallmatrix}
\right)
\left(
\begin{smallmatrix}
Y_1 \\ 
\vdots \\
Y_{18}
\end{smallmatrix}
\right)
=
\left(
\begin{smallmatrix}
0 \\ 
\vdots \\
0
\end{smallmatrix}
\right)
$}
\end{tabular}
$$
Our usual transformation will turn
this into a network flow problem:

$$
\begin{tabular}{c} 
{\rm Minimize} \mbox{$-\sum_{i=1}^{4} l_i Y_{i}+ \sum_{i=1}^4 u_i Y_{4 +i} - \sum_{i=1}^5 a_i Y_{8+i} + \sum_{i=1}^5 b_i Y_{13+i}$} \ \ \ {\rm subject}  \ {\rm to:}\\
\ \\
\mbox{$
\left(
\begin{smallmatrix}
1  & 0  & 0  & 1  &  -1 & 0  & 0  &-1 & 1  & 0  & 0  & 0  & 0  & -1 & 0 & 0 & 0 & 0\\
0  & 1  & 0  & 0  &   0 & -1 & 0  & 0 & -1 & 1  & 0  & 0  & 0  & 1 & -1 & 0 & 0 & 0\\
0  & 0  & 1  & 0  &   0 &  0 & -1 & 0 & 0  & -1 & 1  & 0  & 0  & 0 & 1 & -1 & 0 & 0\\
0  & 0  & 0  & -1 &   0 &  0 &  0 & 1 & 0  & 0  & -1 & 1  & 0  & 0 & 0 & 1 & -1 & 0\\
0  & -1 & 0  & 0  &   0 & 1  &  0 & 0 & 0  & 0  & 0  & -1 & 1  & 0 & 0 & 0 & 1 & -1\\
-1 & 0  & -1 & 0  &   1 & 0  & 1  & 0 & 0  & 0  & 0  & 0  & -1 & 0 & 0 & 0 & 0 & 1
\end{smallmatrix}
\right)
\left(
\begin{smallmatrix}
Y_1 \\ 
\vdots \\
Y_{18}
\end{smallmatrix}
\right)
=
\left(
\begin{smallmatrix}
0 \\ 
\vdots \\
0
\end{smallmatrix}
\right)
$}
\end{tabular}
$$

\begin{figure}
\begin{center}
  \includegraphics[width=0.7\textwidth]{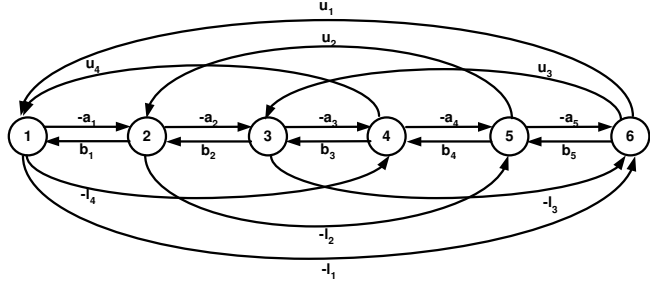}
\end{center}
\caption{Network flow associated to the \SLIDINGSUM\ constraint posted
  on the running example. \label{figure::hardflow}}
\end{figure}

The flow associated with this example
is given in Figure~\ref{figure::hardflow}. There are $n+1$ nodes labelled from 1 to $n+1$
where node $i$ is connected to node $i+1$ with an edge of
cost $-a_i$ and node $i+1$ is connected to node $i$ with an edge of
cost $b_i$. For each window $\myvec{p_i}$, we have an edge from $s_i$
to $s_i + k_i$ with cost $-l_i$ and an edge from $s_i + k_i$ to $s_i$
with cost $u_i$. All nodes have a null supply and a null demand. A
flow is therefore simply a circulation i.e., an amount of flow pushed
on the cycles of the graph.

\begin{theorem}
\label {t:sl_sum}
  The \SLIDINGSUM\ constraint is satisfiable if and only there are no
  negative cycles in the flow graph associated with the dual linear 
  program.
\end{theorem}
\begin{proof}
  If there is a negative cycle in the graph, then we can push an
  infinite amount of flow resulting in a cost infinitely small. Hence
  the dual problem is unbounded, and the primal is
  unsatisfiable. Suppose that there are no negative cycles in the
  graph. Pushing any amount of flow over a cycle of positive cost
  results in a flow of cost greater than zero. Such a flow is not
  optimal since the null flow has a smaller objective value. Pushing any amount of
  flow over a null cycle does not change the objective
  value. Therefore the null flow is an optimal solution and since
  this solution is bounded, then the primal is satisfiable. Note that
  the objective value of the dual (zero) is in this case equal to the
  objective value of the primal. \qed
\end{proof}

\nina{ Based on Theorem~\ref{t:sl_sum} we build  a \emph{BC} filtering algorithm
for the \SLIDINGSUM\ constraint.
The  \SLIDINGSUM\ constraint is \emph{BC} iff the following conditions hold:
\begin{itemize}
	\item value $a_i$ is the lower bound of a variable $X_i$, $i=1,\ldots,n$ iff
	$a_i$ is the smallest value in the domain of $X_i$ such that 
	there are no negative cycles through the edge weighted with $-a_i$ and labeled with the lower
	bound of $X_i$.
	\item value $b_i$ is the upper bound of a variable $X_i$, $i=1,\ldots,n$ iff
	$b_i$ is the greatest value in the domain of $X_i$ such that
	there are no negative cycles through the edge weighted with $b_i$ and labeled with the upper
	bound of $X_i$
	\end{itemize}
}

The flow graph has $O(n)$ nodes and $O(n + m)$ edges. Testing
whether there is a negative cycle takes $O(n^2 + nm)$ time using the
Bellman-Ford algorithm.  
\ignore{We can use this consistency test to construct
an efficient bounds consistency propagator.} We find for each variable
$X_i$ the smallest (largest) value in its domain such that assigning
this value to $X_i$ does not create a negative cycle. We compute the
shortest path between all pairs of nodes using Johnson's algorithm 
in $O(|V|^2 \log |V| + |V||E|)$
time which in our case gives $O(n^2 \log n + n m)$ time.  Suppose
that the shortest path between $i$ and $i+1$ has length $s(i, i+1)$,
then for the constraint to be satisfiable, we need $b_i + s(i,i+1)
\geq 0$. Since $b_i$ is a value potentially taken by $X_i$, we need to
have $X_i \geq - s(i, i+1)$. We therefore assign $\min(\dom(X_i))
\gets \max(\min(\dom(X_i)), -s(i,i+1))$.  Similarly, let the length of
the shortest path between $i+1$ and $i$ be $s(i+1,i)$. For the
constraint to be satisfiable, we need $s(i+1,i) - a_i \geq 0$. Since
$a_i$ is a value potentially taken by $X_i$, we have $X_i \leq
s(i+1,i)$. We assign $\max(X_i) \gets \min(\max(X_i), s(i+1,i))$.  It
is not hard to prove this is sound and complete, removing all values
that cause negative cycles. \xxx[CG]{Maybe but the reviewers might expect a proof anyway.}
Following~\cite{bnqswcp07}, we can make
the propagator incremental using the algorithm by Cotton and
Maler~\cite{cotton-maler-sat06} to maintain the shortest path between $|P|$
pairs of nodes in $O(|E| + |V| \log |V| + |P|)$ time upon edge
reduction. Each time a lower bound $a_i$ is increased or
an upper bound $b_i$ is decreased, the shortest paths can be
recomputed in $O(m + n \log n)$ time.

\ignore{
\section{Cyclic \SEQUENCE\ constraint}

\xxx[MM]{I think we could omit this section, but anyway ive tightened it up}

In rostering problems, we may wish to
produce a cyclic schedule which can
be repeated, say, every four weeks.
We therefore consider a cyclic
version of the \SEQUENCE\ constraint.
More precisely,
$\SEQCYC(l,u,k,[X_1,\ldots,X_n],v)$
ensures that between $l$ and $u$ variables
in $X_i$ to $X_{1+(i+k-2 \mymod n)}$ takes values in the set $v$
for $1 \leq i \leq n$.
This constraint appears to be significantly more difficult than the $\SEQUENCE$ constraint.

For example, the matrix for
$\SEQCYC(l,u,2,[X_1,\ldots,X_3],v)$
contains, as a submatrix,
\begin{eqnarray*}
\left( \begin{array}{ccc}
1 & 1 & 0\\
0 & 1 & 1\\
1 & 0 & 1
\end{array} \right)
\end{eqnarray*}
\noindent
which has determinant 2.
Thus the matrix is not totally unimodular.

\begin{proposition}
In general, the \SEQCYC\ constraint does not have a totally unimodular matrix.
\end{proposition}

This means, in particular,
that methods employing a network flow or its dual,
as in Sections~\ref{sect:flowseq} and \ref{sect:genseq},
are not directly applicable.
Even the use of linear programming,
as discussed in Section~\ref{sect:genseq} for $\SEQUENCE$,
is not available.

\ignore{
\begin{theorem}
In general, the \SEQCYC\ constraint does not have a totally unimodular matrix.
The constraint matrix of the LP associated
with the cyclic version of the \SEQUENCE\ constraint might not
 be totally unimodular.
\end{theorem}
\begin{proof}
 A cyclic \SEQUENCE\ constraint over 3 variables with
  a window of size 2 can be written as follows:

$$
\left(
\begin{smallmatrix}
1 & 1 & 0 & 1 & 0 & 0 & 0 & 0 & 0 \\
1 & 1 & 0 & 0 & -1& 0 & 0 & 0 & 0 \\
0 & 1 & 1 & 0 & 0 & 1 & 0 & 0 & 0 \\
0 & 1 & 1 & 0 & 0 & 0 & -1& 0 & 0 \\
1 & 0 & 1 & 0 & 0 & 0 & 0 & 1 & 0 \\
1 & 0 & 1 & 0 & 0 & 0 & 0 & 0 & -1
\end{smallmatrix}
\right)
\left(
\begin{smallmatrix}
X_1\\
X_2\\
X_3\\
Y_1\\
Z_1\\
Y_2\\
Z_2\\
Y_3\\
Z_3
\end{smallmatrix}
\right)
\geq
\left(
\begin{smallmatrix}
l\\
u\\
l\\
u\\
l\\
u
\end{smallmatrix}
\right)
$$

For the constraint matrix to be totally
unimodular, all its square sub-matrices must have a determinant
included in $\{-1, 0, 1\}$. This is not the case for the following
sub-matrix.

\begin{eqnarray*}
\left| \begin{array}{ccc}
1 & 1 & 0\\
0 & 1 & 1\\
1 & 0 & 1
\end{array} \right| & = & 2
\end{eqnarray*}
\qed
\end{proof}

Since the constraint 
matrix of the LP associated with the cyclic \SEQUENCE\ may
not be totally unimodular, we 
cannot necessarily represent it by a network flow.
Thus, we cannot use the same flow based
methods proposed in the rest of this paper
to propagate cyclic forms of the \SEQUENCE\ constraint. 
}
}

\section {Experimental Results}    \label{sect:expt}
To evaluate the performance of our filtering algorithms we carried out a series 
of experiments on random problems. The experimental setup is similar
to that in \cite{bnqswcp07}.
The first set of experiments compares performance of the flow-based propagator $\fl$
on single instance of the $\SEQUENCE$ constraint against the $\hprs$ propagator\footnote{We would like to thank Willem-Jan van Hoeve for providing us with the implementation of the $\hprs$
algorithm.}
(the third propagator in \cite{Hoeve06}), 
the $\cs$ encoding of \cite{bnqswcp07}, 
and the $\AMONG$ decomposition ($\among$) of $\SEQUENCE$.
The second set of experiments compares the flow-based propagator $\flS$ for the 
$\SOFTSEQ$ constraint and its decomposition into soft $\AMONG$ constraints. Experiments were run with ILOG 6.1 on an
Intel Xeon 4 CPU, 2.0 Ghz, 4G RAM. Boost graph library version $1.34.1$ 
was used to implement the flow-based algorithms.   

\vspace*{-15pt}
\begin{figure} \centering
\begin{tabular}{cc} \centering
  \includegraphics[width=0.52\textwidth]{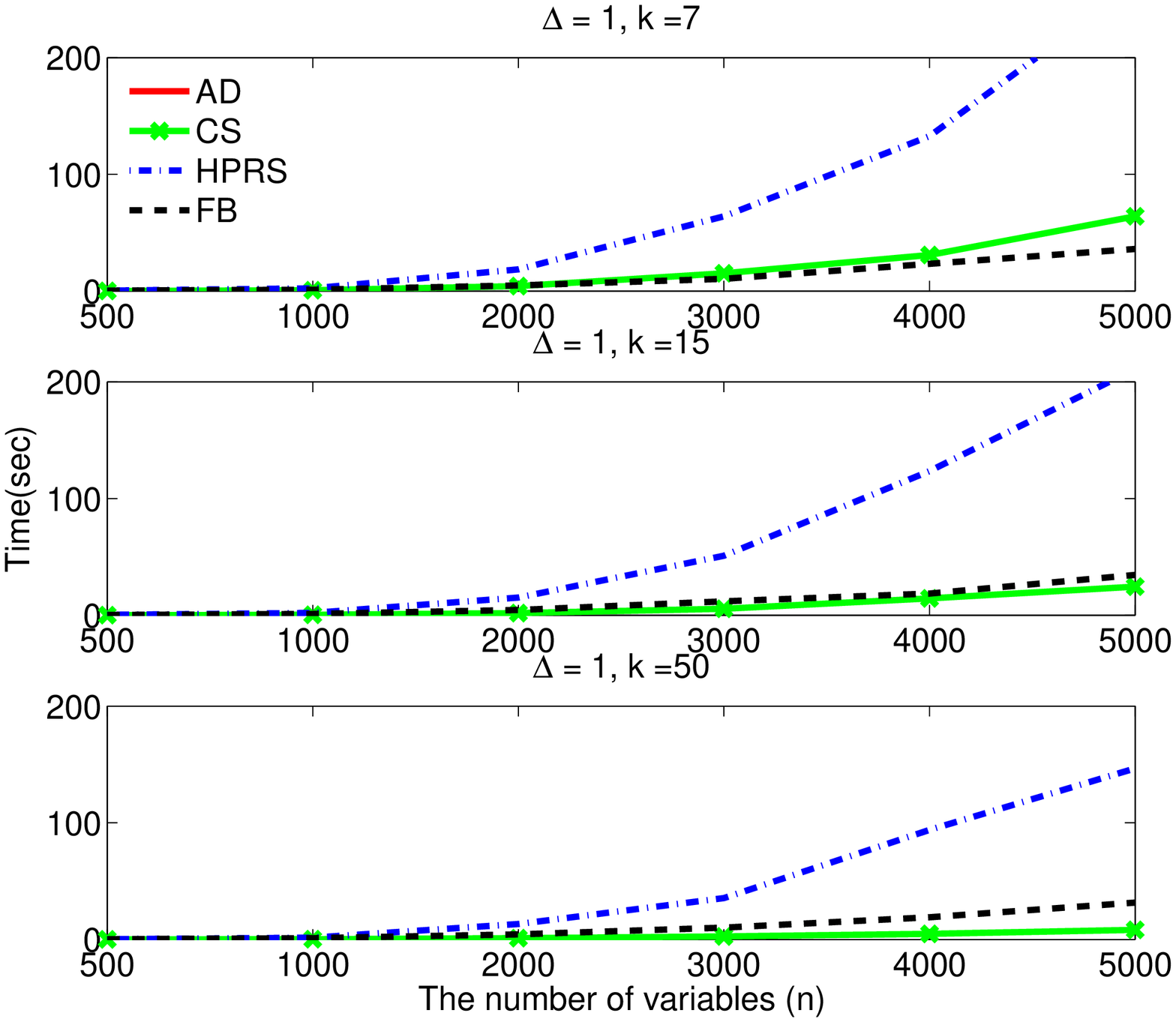} &   \includegraphics[width=0.52\textwidth]{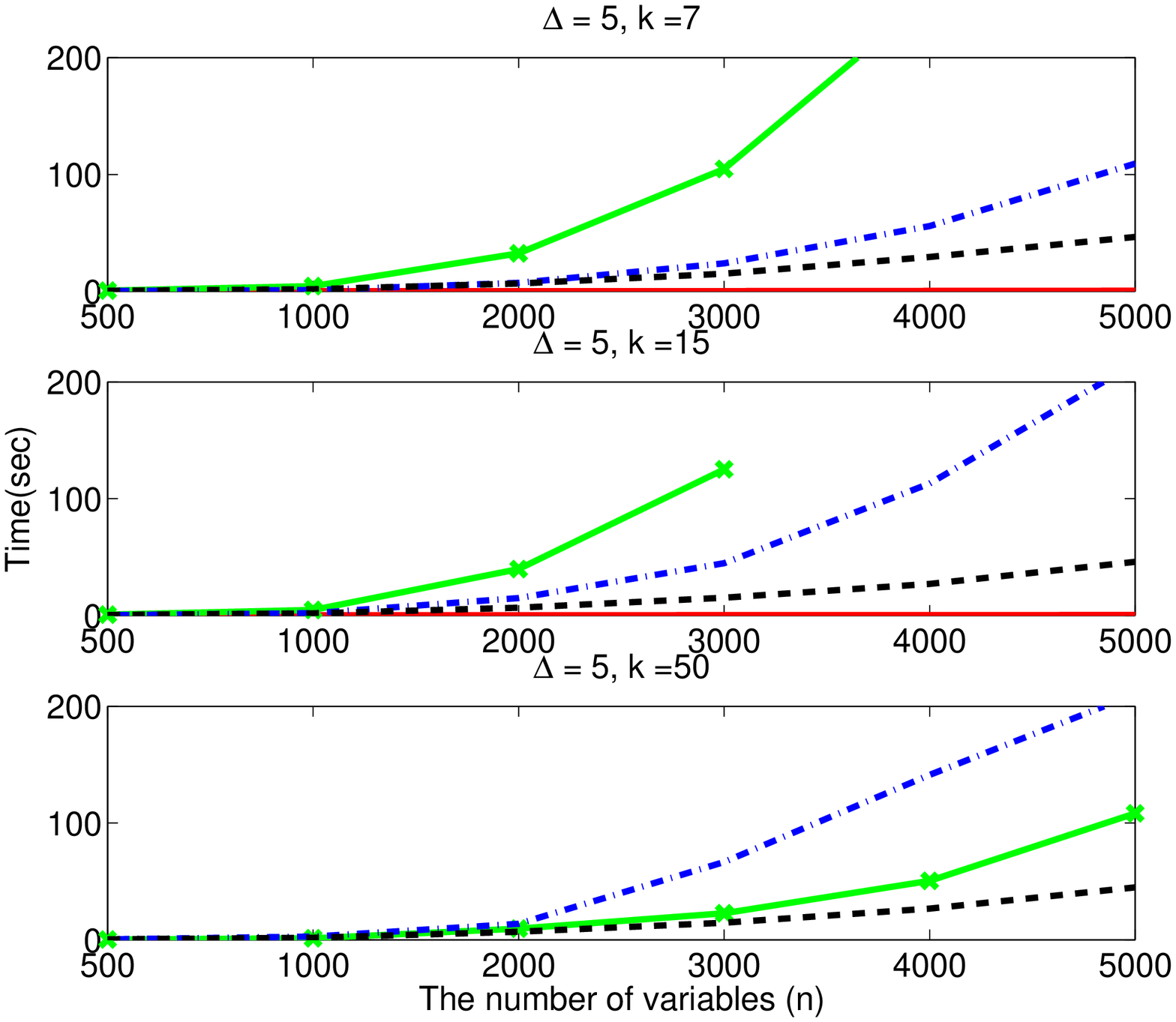} \\ 
\end{tabular}
\scriptsize {\caption{\label{f:deltas} Randomly generated instances with a single $\SEQUENCE$ constraints for different combinations of $\Delta$ and $k$.}
}
\end{figure}

\vspace*{-30pt}
\subsection{The $\SEQUENCE$ constraint}

For each possible combination of $n\in \{500,1000,2000,3000,4000,5000\}$,
$k \in \{5,15,50\}$, $\Delta=u-l \in \{1,5\}$, we generated twenty  
instances with random lower bounds in the interval $(0,k-\Delta)$. We used
random value and variable ordering and a time out of $300$ sec.
We used the Ford-Fulkerson algorithm to find a  maximum flow.  Results for different values of $\Delta$ are presented in 
Tables~\ref{t:t1},~\ref{t:t2} and Figure~\ref{f:deltas}.
\nina{Table~\ref{t:t1} shows results for tight problems with $\Delta = 1$ 
and Table~\ref{t:t2} for easy problems with $\Delta = 5$.
To investiage empirically the asymptotic growth 
of the different propagators, 
we plot average time to solve 20 instances against the instance size 
for each combination of parameters $k$ and $\Delta$ in Figure~\ref{f:deltas}.}
First of all, we notice 
that the $\cs$ encoding is the best on hard instances ($\Delta=1$) and the $\among$ decomposition is the
fastest on easy instances ($\Delta=5$). This result was first observed in ~\cite{bnqswcp07}.
The $\fl$ propagator is not the fastest one but has the most robust 
performance.
It is sensitive only to the value of $n$ and not to other parameters, like the 
length of the window($k$) or hardness of the problem($\Delta$). 
As can be seen from Figure~\ref{f:deltas}, the $\fl$ propagator scales better than the
other propagators with the size of the problem. It appears to grow linearly with the number of variables, 
while the $\hprs$ propagator display quadratic growth.

\vspace*{-12pt}
\begin{table}
\begin{center}
{\scriptsize
\caption{\label{t:t1}Randomly generated instances with a single $\SEQUENCE$ constraint and $\Delta=1$.
Number of instances solved in 300 sec / average time to solve.
We omit results for $n\in\{1000, 3000, 4000\}$ due to space limitation. The summary rows include all instances.
}
\begin{tabular}{|  c c |cc|cc|cc| cc|cc|}
\hline
$n$ & $k$
&\multicolumn {2}{|c|}{$\among$}
&\multicolumn {2}{|c|}{$\cs$}
&\multicolumn {2}{|c|}{$\hprs$}
&\multicolumn {2}{|c|}{$\fl$} \\
\hline 
\hline 
 {500} &  {7}   & 8 & /   2.13 & \textbf{20}& / \textbf{   0.13} & \textbf{20}& /   0.35 & \textbf{20}& /   0.30 \\ 
 & {15}   & 6 & /   0.01 & \textbf{20}& / \textbf{   0.09} & \textbf{20}& /   0.30 & \textbf{20}& /   0.29 \\ 
 & {50}   & 2 & /   0.02 & \textbf{20}& / \textbf{   0.07} & \textbf{20}& /   0.26 & \textbf{20}& /   0.28 \\ 
\hline 
\hline 
 {2000} &  {7}   & 4 & /   0.04 & \textbf{20}& / \textbf{   4.25} & \textbf{20}& /  18.52 & \textbf{20}& /   4.76 \\ 
 & {15}   & 0 & /0 & \textbf{20}& / \textbf{   1.84} & \textbf{20}& /  15.19 & \textbf{20}& /   4.56 \\ 
 & {50}   & 1 & /0 & \textbf{20}& / \textbf{   1.16} & \textbf{20}& /  13.24 & \textbf{20}& /   4.42 \\ 
\hline 
\hline 
 {5000} &  {7}   & 1 & /0 & \textbf{20}& /  64.05 & 15 & / 262.17 & \textbf{20}& / \textbf{  36.09} \\ 
 & {15}   & 0 & /0 & \textbf{20}& / \textbf{  24.46} & 17 & / 211.17 & \textbf{20}& /  34.59 \\ 
 & {50}   & 0 & /0 & \textbf{20}& / \textbf{   8.24} & 19 & / 146.63 & \textbf{20}& /  31.66 \\ 
\hline 
\hline 
\multicolumn {2}{|r|} { TOTALS }& & & & & & & & \\ 
\multicolumn {2}{|r|} {solved/total}& 37 &/360& \textbf{360}& /360& 351 &/360& \textbf{360}& /360\\ 
\multicolumn {2}{|r|} {avg time for solved}& \multicolumn {2}{|c|}{  0.517} & \multicolumn {2}{|c|}{\textbf{  9.943}} & \multicolumn {2}{|c|}{ 60.973} & \multicolumn {2}{|c|}{ 11.874} \\ 
\multicolumn {2}{|r|} {avg bt for solved}& \multicolumn {2}{|c|}{  17761} & \multicolumn {2}{|c|}{    429} &\multicolumn {2}{|c|}{0} &\multicolumn {2}{|c|}{0} \\ 
\hline 
\end{tabular}
}
\end{center}
\end{table}

\vspace*{-40pt}
\begin{table}
\begin{center}
{\scriptsize
\caption{\label{t:t2}Randomly generated instances with a single $\SEQUENCE$ constraint and $\Delta=5$.
Number of instances solved in 300 sec / average time to solve. 
We omit results for $n\in\{1000, 3000, 4000\}$ due to space limitation. The summary rows include all instances.
}
\begin{tabular}{|  c c |cc|cc|cc| cc|cc|}
\hline
$n$ & $k$
&\multicolumn {2}{|c|}{$\among$}
&\multicolumn {2}{|c|}{$\cs$}
&\multicolumn {2}{|c|}{$\hprs$}
&\multicolumn {2}{|c|}{$\fl$} \\
\hline 
\hline 
 {500} &  {7}   & \textbf{20}& / \textbf{   0.01} & \textbf{20}& /   0.58 & \textbf{20}& /   0.15 & \textbf{20}& /   0.44 \\ 
 & {15}   & \textbf{20}& / \textbf{   0.01} & \textbf{20}& /   0.69 & \textbf{20}& /   0.25 & \textbf{20}& /   0.44 \\ 
 & {50}   & 18 & /   0.02 & \textbf{20}& / \textbf{   0.20} & \textbf{20}& /   0.37 & \textbf{20}& /   0.42 \\ 
\hline 
\hline 
 {2000} &  {7}   & \textbf{20}& / \textbf{   0.07} & \textbf{20}& /  32.41 & \textbf{20}& /   7.19 & \textbf{20}& /   6.62 \\ 
 & {15}   & \textbf{20}& / \textbf{   0.07} & \textbf{20}& /  39.71 & \textbf{20}& /  14.89 & \textbf{20}& /   6.63 \\ 
 & {50}   & 5 & /   5.19 & \textbf{20}& /   9.52 & \textbf{20}& /  13.71 & \textbf{20}& / \textbf{   6.94} \\ 
\hline 
\hline 
 {5000} &  {7}   & \textbf{20}& / \textbf{   0.36} & 0 & /0 & \textbf{20}& / 109.18 & \textbf{20}& /  46.42 \\ 
 & {15}   & \textbf{20}& / \textbf{   0.36} & 6 & / 160.99 & 17 & / 215.97 & \textbf{20}& /  45.97 \\ 
 & {50}   & 9 & /   0.48 & \textbf{20}& / 108.34 & 11 & / 210.53 & \textbf{20}& / \textbf{  44.88} \\ 
\hline 
\hline 
\multicolumn {2}{|r|} { TOTALS }& & & & & & & & \\ 
\multicolumn {2}{|r|} {solved/total}& 296 &/360& 308 &/360& 345 &/360& \textbf{360}& /360\\ 
\multicolumn {2}{|r|} {avg time for solved}& \multicolumn {2}{|c|}{  0.236} & \multicolumn {2}{|c|}{ 52.708} & \multicolumn {2}{|c|}{ 50.698} & \multicolumn {2}{|c|}{\textbf{ 16.200}} \\ 
\multicolumn {2}{|r|} {avg bt for solved}& \multicolumn {2}{|c|}{    888} & \multicolumn {2}{|c|}{   1053} &\multicolumn {2}{|c|}{0} &\multicolumn {2}{|c|}{\textbf{0}} \\ 
\hline 
\end{tabular}
}
\end{center}
\end{table}

\vspace*{-20pt}
\subsection{The Soft $\SEQUENCE$ constraint} 

We evaluated performance of the soft $\SEQUENCE$ constraint on random problems.
For
each possible combination of   $n\in\{50,100\}$, $k\in\{5,15,25\}$,
$\Delta=\{1,5\}$ and $m\in\{4\}$ (where $m$ is the number of $\SEQUENCE$
constraints), we generated twenty random instances.
All variables had domains of size 5.
An instance was obtained by selecting random lower bounds
in the interval  $(0,k-\Delta)$.
We excluded instances where $\sum_{i=1}^m l_i \geq k$ to
avoid unsatisfiable instances. We used
a random variable and value ordering, and a time-out of $300$ sec.
All $\SEQUENCE$ constraints were enforced on disjoint sets of cardinality one.
\nina {Instances with  $\Delta=1$ are  hard instances for $\SEQUENCE$
propagators~\cite{bnqswcp07}, so that any \emph{DC} propagator could  solve
only few instances. Instances with $\Delta = 5$ are much looser problems, but they are
still hard do solve because each instance includes four overlapping $\SEQUENCE$ constraints}.
To relax these instances, we
allow the $\SEQUENCE$ constraint to be violated with a cost that has to be less than or equal to $15\%$ of the length of 
the sequence.
Experimental results are presented in Table~\ref{t:multi_t1}.
As can be seen from the table, the  $\flS$ algorithms is competitive with 
the decomposition into soft $\AMONG$ constraints on \nina{relatively} easy problems
and outperforms the decomposition on hard problems \nina{in terms of the number of solved problems}. 
 
We observed that the flow-based propagator for the $\SOFTSEQ$ constraint ($\flS$) is very slow. 
Note that the number of backtracks of 
$\flS$ is three order of magnitudes smaller compared to $\amongS$.
We profiled the algorithm and found that it spends
most of the time performing the all pairs shortest path algorithm.
Unfortunately, this is difficult to compute
incrementally because the residual graph can be different on every
invocation of the propagator.

\vspace*{-12pt}
\begin{table}
\begin{center}
{\scriptsize
\caption{\label{t:multi_t1}Randomly generated instances with 4 soft $\SEQUENCE$s.
Number of instances solved in 300 sec / average time to solve.
}
\begin{tabular}{|  c c |cc|cc|cc|cc|cc|}
\hline
& &  \multicolumn {4}{|c|}{$\Delta=1$} & \multicolumn {4}{|c|}{$\Delta=5$} \\
\hline
$n$ & $k$
&\multicolumn {2}{|c|}{$\amongS$}
&\multicolumn {2}{|c|}{$\flS$} 
&\multicolumn {2}{|c|}{$\amongS$}
&\multicolumn {2}{|c|}{$\flS$} 
\\
\hline 
\hline 
 {50} &  {7}   & 6 & /  19.30 & \textbf{7}& / \textbf{  27.91} & \textbf{20}& / \textbf{   0.01} & \textbf{20}& /   2.17 \\ 
 & {15}   & 8 & /  36.07 & \textbf{13}& / \textbf{  20.41} & \textbf{11}& / \textbf{  49.49} & 10 & /  30.51 \\ 
 & {25}   & 6 & /   0.73 & \textbf{10}& / \textbf{  23.27} &  \textbf{10}& / \textbf{   6.40} & \textbf{10}& /   7.41 \\ 
\hline 
\hline 
 {100} &  {7}   & 1 & /0 & \textbf{3}& / \textbf{   7.56} &  \textbf{19}& / \textbf{  10.50} & 18 & /  16.51 \\ 
 & {15}   & 0 & /0 & \textbf{5}& / \textbf{   6.90} & \textbf{3}& / \textbf{   0.01} & \textbf{3}& /   7.20 \\ 
 & {25}   & 0 & /0 & \textbf{5}& / \textbf{   4.96} &  \textbf{5}& / \textbf{  19.07} & \textbf{5}& /  23.99 \\ 
\hline 
\hline 
\multicolumn {2}{|r|} { TOTALS }& & & & & & & & \\ 
\multicolumn {2}{|r|} {solved/total}& 21 &/120& \textbf{43}& /120 &  \textbf{68}& /120& 66 &/120 \\  
\multicolumn {2}{|r|} {avg time for solved}& \multicolumn {2}{|c|}{ 19.463} & \multicolumn {2}{|c|}{\textbf{ 18.034}} & \multicolumn {2}{|c|}{\textbf{ 13.286}} & \multicolumn {2}{|c|}{ 13.051} \\ 
\multicolumn {2}{|r|} {avg bt for solved}& \multicolumn {2}{|c|}{ 245245} &\multicolumn {2}{|c|}{\textbf{    343}} & \multicolumn {2}{|c|}{\textbf{ 147434}} & \multicolumn {2}{|c|}{    128} \\ 
\hline 
\end{tabular}
}
\end{center}
\end{table}

\vspace*{-30pt}
\section {Conclusion}

We have proposed new filtering algorithms for the \SEQUENCE\ constraint
and several extensions including the soft \SEQUENCE\ and generalized
\SEQUENCE\ constraints which are based on
network flows. Our propagator
for the \SEQUENCE\ constraint enforces domain consistency in $O(n^2)$
time down a branch of the search tree. 
This improves upon the best existing
domain consistency algorithm by a factor of $O(\log n)$.
We also introduced a soft version of the \SEQUENCE\ constraint 
and 
{propose an $O(n^2\log n \log\log u)$ time}
domain consistency algorithm
based on minimum cost network flows. 
These algorithms
are derived from linear programs which
represent a network flow. 
They differ from the flows used to propagate global
constraints like \gcc\ since the domains
of the variables are encoded as costs 
on the edges rather than capacities. 
Such flows
are efficient for maintaining bounds consistency over large domains.
{Experimental results demonstrate that the $\fl$
filtering algorithm is more robust than
existing propagators.}
We conjecture that similar flow
based propagators derived from linear
programs may be useful for other global
constraints. 


\bibliographystyle{splncs}



\end{document}